\documentclass{article}

\usepackage[preprint]{neurips_2025}

\usepackage[utf8]{inputenc} 
\usepackage[T1]{fontenc}    
\usepackage{hyperref}       
\usepackage{url}            
\usepackage{booktabs}       
\usepackage{amsfonts}       
\usepackage{nicefrac}       
\usepackage{microtype}      
\usepackage{lipsum}
\usepackage{graphicx}
\usepackage{subcaption}
\usepackage{algorithm}
\usepackage{algpseudocode}
\usepackage{placeins}

\usepackage{wrapfig}

\usepackage{amsfonts, amsmath,amssymb, amsthm}
    \newtheoremstyle{spacedstyle} 
      {10pt} 
      {10pt} 
      {\itshape} 
      {} 
      {\bfseries} 
      {.} 
      { } 
      {} 

    \theoremstyle{spacedstyle}

    \newtheorem{theorem}{Theorem}[section] 

    \theoremstyle{definition}
    \newtheorem{definition}[theorem]{Definition}

    \theoremstyle{remark}
    
\usepackage{xcolor}
\graphicspath{ {./images/} }

\title{In-Context Curiosity: Distilling Exploration for Decision-Pretrained Transformers on Bandit Tasks}

\author{
    Huitao Yang\\
  University of California, Los Angeles\\
  \texttt{htyang03@ucla.edu} \\
   \And
   Guanting Chen\\
   University of North Carolina at Chapel Hill\\
     \texttt{guanting@unc.edu} \\
}

\begin{document}
\maketitle
\begin{abstract}
As large language models (LLMs) continue to grow in capability, there is increasing interest in incorporating them into decision-making tasks. A common pipeline for this is \textit{Decision-Pretrained Transformers} (DPTs). However, existing training methods for DPTs often struggle to generalize beyond their pretraining data distribution. To explore mitigation of this limitation, we propose \textit{in-context curiosity}---a lightweight, exploration-inspired regularizer for offline pretraining---and introduce the \textit{Prediction-Powered Transformer (PPT)} framework. PPT augments DPT with an auxiliary reward predictor, using prediction error as an intrinsic curiosity signal to encourage broader exploration during training. In proof-of-concept experiments on Gaussian multi-armed bandits, PPT shows improved robustness: it moderates the performance degradation observed in DPT when test environments exhibit higher variance in reward, particularly when pretraining data has limited diversity. While the quality of offline data remain fundamental, our preliminary results suggest that curiosity-driven pretraining offers a promising direction for enhancing out-of-distribution generalization in in-context RL agents. 
\end{abstract}


\section{Introduction}

In-context reinforcement learning (RL) has recently emerged as a versatile paradigm for leveraging pre-collected datasets to train agents that can generalize to new environments. A series of works on sequence-model-based agents~\citep{dt, laskin2022incontextreinforcementlearningalgorithm, xu2025tackling} demonstrate that pretrained transformers can learn complex RL policies directly from offline trajectories, without requiring online interaction. These approaches highlight the potential of in-context learning as a general-purpose framework for decision-making. Among them, Decision-Pretrained Transformers (DPT)~\citep{dpt} stand out for their simple training pipeline: by directly optimizing negative log-likelihood on offline trajectories, they extract generalizable decision-making patterns from diverse pretraining tasks. This simplicity has made DPT an attractive foundation, leading to a number of recent variants and extensions~\citep{mukherjee2025pretrainingdecisiontransformersreward, rietz2025enhancingpretraineddecisiontransformers, hu2024qvalueregularizedtransformeroffline}. However, a key limitation remains: DPTs generalize well only when trained on diverse, exploratory datasets. With biased data, they tend to pick up spurious correlations—performing strongly in-distribution but failing to transfer to out-of-distribution (OOD) settings, even in the simple bandit setting \cite{lin2024transformers,wang2025understanding,goddard2025incontextlearninggeneralizetask}.

In this work, we study this deficiency in the simplest multi-armed bandit (MAB) setting. Unlike in sequential tasks, where replicating parts of the memory can sometimes suffice, success in bandits depends on robust generalization beyond dataset-specific patterns. Motivated by the idea of curiosity in online RL~\citep{pathak2017curiositydrivenexplorationselfsupervisedprediction}, we propose \textit{in-context curiosity}, an exploration-driven regularizer that operates during pretraining. Unlike classical intrinsic reward methods that rely on online rollouts, our curiosity term is incorporated directly into the offline training objective. We then embed this mechanism into the DPT pipeline, leading to the Prediction-Powered Transformer (PPT) framework (Section~\ref{sec:ppt}), which introduces an auxiliary predictor to quantify uncertainty and compute curiosity.

In Section~\ref{sec:exp}, we empirically show that in-context curiosity improves generalization and robustness in MABs by narrowing the performance gap between in-distribution and OOD environments. Nevertheless, the limitation of offline data remains the primary obstacle: while curiosity regularization reduces bias, it cannot fully replace exploration. In particular, our method should not be seen as a complete solution; the principled way to address distribution shift is to model posterior uncertainty at each time step, as in Bayesian sampling, which remains an important direction for future work.

\textbf{Contribution}. Our main contributions are: 
\begin{itemize}
\item We propose to incorporate curiosity-driven exploration into the pre-training of DPT. The method requires only minor changes, preserves training simplicity, and can be applied to in-context RL tasks.  
\item Our exploration-based pre-training mitigates the lack of exploration in standard DPT performance. In controlled experiments on Gaussian bandits, it reduces the performance degradation observed in DPT, particularly when the variance of test environments increases.  
\end{itemize}
\section{In-Context Curiosity}\label{sec:cur}

\begin{figure}[t]
      \begin{subfigure}[b]{0.45\textwidth}
        \centering
    \includegraphics[width=\textwidth]{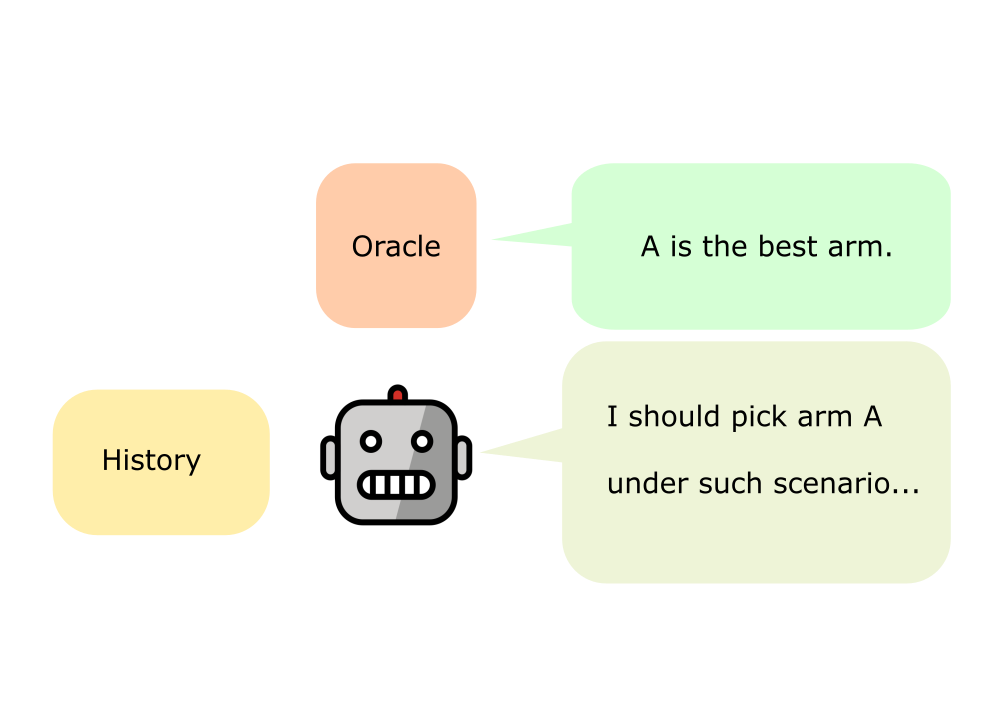}
        \caption{DPT}
      \end{subfigure} 
    \hfill
          \begin{subfigure}[b]{0.45\textwidth}
        \centering
    \includegraphics[width=\textwidth]{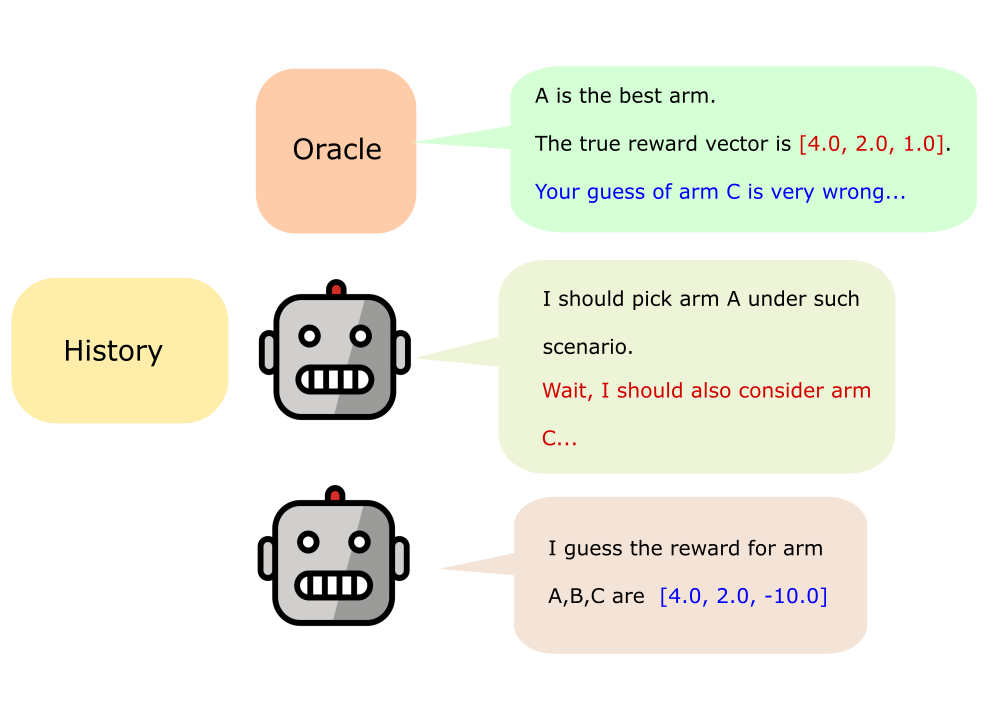}
        \caption{Pretrain with in-context curiosity}
      \end{subfigure} 
    \caption{An illustrative diagram of in-context curiosity during pretraining. An additional round of prediction and self-reflection is incorporated to encourage exploration.}
    \label{fig:cur}
\end{figure}



\paragraph{Decision-Pretrained Transformers.}
In the standard pipeline of DPT, training starts with the collection of trajectories obtained from interactions with sampled environments under some policies. A transformer is then trained via supervised learning on this offline dataset $\mathcal{B}$. The pretraining objective \eqref{ldpt} aims to predict the optimal action $a^*$ given $D_j = \{o_1,a_1,r_1,\cdots, o_{j-1}, a_{j-1}, r_{j-1}, o_j\}$, the ``current'' observation of states, actions, and rewards at each time step $j$.

\begin{equation}\label{ldpt}
\mathcal{B}  \sim  \mathcal{T}_{\text{pre}}, ~~~
     \min_{\theta} L^{\texttt{DPT}}_{\theta} = \mathbb{E}_{\mathcal{T}_{\text{pre}}}\sum_{j \in [n]} 
        -\log \pi_\theta(a^* \mid D_j)
\end{equation}

At deployment, the model observe the current trajectory, make decision by sampling from the policy of the learned Transformer, and get new reward and observation by interacting with the environment. When applied in a online setting, this mechanism reveals several limitations. The key constraint is that the pre-collected dataset is the only “teacher” available, and the resulted transformer lack exploration. If an online observation $\hat D_j$ falls on the fringes of the training distribution, where coverage is poor, the knowledge distilled during pretraining may fail to guide the model toward the correct arm. Worse still, if the dataset is too expert-biased (the extreme case is that the data-collection policy always selects the best arm), the model may learn misleading lessons and exploit the wrong action. Such collapse is especially likely when data is dominated by expert demonstrations or exhibits sparse exploration, and when online environments have high-variance rewards.

\begin{align}
    \texttt{limited or biased data} \Rightarrow \texttt{uncertain or unreliable prediction }   \hat D_j \rightarrow \hat a^*. 
\end{align}

To mitigate this collapse, a natural idea is to bias the model toward \textit{exploring when necessary}. Encouraging exploration helps avoid premature commitment to misleading arms and can steer the online trajectory back into regions well-covered by pretraining data. This raises the central question: \textit{Can exploration be distilled during pretraining?} Towards this goal, we are motivated by the concept of curiosity-driven exploration and implant an in-context version of curiosity into the pretraining objective.

\paragraph{Curiosity-driven Exploration.}
Curiosity as an exploration signal is well-established in reinforcement learning, particularly in online settings where intrinsic rewards are added to the observed rewards to encourage visiting uncertain states \citep{pathak2017curiositydrivenexplorationselfsupervisedprediction, burda2018explorationrandomnetworkdistillation}. In the Intrinsic Curiosity Module (ICM) of \citep{pathak2017curiositydrivenexplorationselfsupervisedprediction}, curiosity is defined via a forward dynamics model: given a state--action pair $(s_t, a_t)$, a learned predictor $f_\phi$ estimates the next state’s feature embedding, and the intrinsic reward is given by
\begin{equation}
r^{\text{int}}_t = \| f_\phi(s_t, a_t) - \phi(s_{t+1}) \|^2.
\end{equation}
The agent is thus incentivized to visit transitions that are hard to predict, and this intrinsic term augments the environment reward during rollouts.

In our setting of pretrained decision models, we depart from this online framework: pretraining is \emph{fully offline} on a fixed dataset collected prior to optimization. This prohibits augmenting the reward stream directly. Instead, we design a \emph{curiosity term} that integrates into the \emph{pretraining objective} itself. The goal is to bias the policy toward actions whose outcomes are under-predicted or uncertain according to a learned predictor, thereby shaping the model’s in-context behavior without modifying the dataset. An illustration for the motivation is Figure~\ref{fig:cur}, where we apply the pretraining data (oracle) and an extra step of prediction for self-supervised training.

\paragraph{Application in the Bandit Setting.} While prior work formulates curiosity through prediction errors on \emph{state transitions}, the multi-armed bandit setting has no state dynamics to learn. The natural analogue is to use a \emph{reward predictor}, which plays the role of a Q-function learner in reinforcement learning. Specifically, our predictor estimates the mean reward of each arm, and the curiosity signal is obtained from the squared error between predicted and ground-truth mean rewards. In simulation, the ground-truth can be accessed directly; in practice, it may be replaced by an empirical estimator derived from offline data. Thus, our construction mirrors the spirit of the original ICM formulation but specializes it to the bandit case, where curiosity reduces to reward-prediction error.

Formally, we define curiosity for an action $a$ as the squared error between its true expected reward and the predictor’s estimate $\hat\mu(a)$:
\begin{equation}
\texttt{curiosity} (a): = \|\mathbb{E} R(a) - \hat{\mu}(a) \|^2,
\end{equation}
where $\hat\mu(a)$ denotes the predictor’s reward estimate for action $a$. During training, we aim to teach the model to select arms of high curiosity with a curiosity-maximizing function (see~\ref{lppt}).

This formulation is naturally compatible with the multi-armed bandit setting and integrates cleanly with DPT training pipelines. The formal construction of the curiosity term and its integration into the full objective are detailed in Section~\ref{sec:ppt}.
\section{Prediction-Pretrained Transformers}\label{sec:ppt}

\FloatBarrier
\begin{algorithm}[t]
\caption{Prediction Powered Transformer (PPT): Training and Deployment}
\label{algo:ppt}
\begin{algorithmic}[1]
\Statex \textcolor{blue}{//~~Collecting pretraining dataset}
\State Initialize empty pretraining dataset $\mathcal{B}$
\For{$i$ in $[N]$}
    \State Sample environment $\tau \sim \mathcal{T}_{\text{pre}}$ and dataset $D \sim \mathcal{D}_{\text{pre}}(\cdot | \tau)$ 
    \State Add $(D, a^*, c^*)$ to $\mathcal{B}$
\EndFor
\Statex \textcolor{blue}{//~~Pretraining model on dataset}
\State Initialize policy model $\pi_{\theta}$ and predictor model $q_{\phi}$
\While{not converge}
    \State \textcolor{blue}{//~~Rolling out predictions and update predictor model}
    \State Sample $(D, a^*. c^*)$ from $ \mathcal{B}$ and predict $c_j = q_{\phi}(\cdot | D_j)$ for $j \in [n]$
    \State Compute loss in \ref{lpred} and backpropagate to update $q_{\phi}$ 
    \State \textcolor{blue}{//~~Rolling out actions  and update policy model}
    \State Predict $p_j = \pi_{\theta}(\cdot | D_j, c_{1,2,\ldots, j})$ for $j \in [n]$
    \State Compute loss in \ref{lppt} and backpropagate to update $\pi_{\theta}$
\EndWhile

\Statex \textcolor{blue}{//~~Online test-time deployment}
\State Sample Environment $\tau \sim \mathcal{T}_{\text{test}}$ and initialize empty $D = \{\}$ and $c =  \{\}$
\For{$j \in [n_{\text{test}}]$}
\State Deploy $q_{\phi}$ to predict $c_j = q_{\phi}(\cdot | D)$ and add $c_j$ to $c$ 
\State Deploy $\pi_{\theta}$ to sample $a_j \sim \pi_{\theta}(\cdot | D, c)$ and add $(a_j, r_j)$ to $D$
\EndFor
\end{algorithmic}
\end{algorithm}

\paragraph{Model architecture.} 
Our framework augments the standard DPT with an auxiliary predictor model. Specifically, we employ two components: an autoregressive transformer $\pi_\theta$, which plays the role of the \emph{policy model}, and a sequential predictor $q_\phi$ (e.g., another transformer of comparable scale). At each interaction round $j$, the predictor outputs a reward estimate $c_j$, which is appended to the history and passed as an additional input to the policy. In comparison to DPT, this modification allows the policy to condition not only on past actions and rewards, but also on predicted outcomes, enabling curiosity-driven exploration.

\paragraph{Training and deployment.}
Training is very natural based on the current DPT pipeline. The models are trained on a pre-collected dataset $\mathcal{B}$, obtained by sampling environments $\tau \sim \mathcal{T}_{\text{pre}}$ and generating trajectories $D \sim \mathcal{D}_{\text{pre}}(\cdot|\tau)$ using a fixed random policy. For each environment, the optimal action $a^*$ and the true expected reward vector $c^*$ are collected:
\begin{equation}
    a^* = \arg\max_{a \in \mathcal{A}} \mathbb{E}R(a), 
    \qquad 
    c^* = \big(\mathbb{E}R(a_1), \ldots, \mathbb{E}R(a_{|\mathcal{A}|})\big) \in \mathbb{R}^{|\mathcal{A}|}.
\end{equation}

Substituting the ground-truth value $c^*$ with an empirical estimate, such as the mean reward observed over the full episode, provides a feasible alternative that does not rely on privileged access to the true reward function (see \ref{sec:ctx_prox}).
The predictor $q_\phi$ is trained to regress toward $c^*$, minimizing the mean-squared error along each trajectory:
\begin{equation}\label{lpred}
    \min_{\phi} L_{\phi} = \mathbb{E}\sum_{j \in [n]} \| q_\phi(\cdot | D_j) - c^* \|_2^2.
\end{equation}
The policy $\pi_\theta$ is updated via the standard negative log-likelihood (NLL) loss, augmented with a weighted curiosity term:
\begin{equation}\label{lppt}
    \min_{\theta} L_{\theta} = \mathbb{E}_{\mathcal{T}_{\text{pre}}}\sum_{j \in [n]} 
    \Big[ 
        \underbrace{-\log \pi_\theta(a^* \mid D_j, c_{1{:}j})}_{\textcolor{blue}{\text{NLL loss}}} 
        - \lambda \cdot 
        \underbrace{\langle \mathcal{E}_j, \pi_\theta(\cdot \mid D_j, c_{1{:}j}) \rangle}_{\textcolor{green}{\text{curiosity bonus}}}
    \Big],
\end{equation}
where the curiosity vector is defined as the element-wise squared error between predicted and true mean rewards:
\begin{equation}
    \mathcal{E}_j = \big(q_\phi(\cdot|D_j) - c^*\big)^{\odot 2} \in \mathbb{R}^{|\mathcal{A}|}.
\end{equation}

At test time, environments are drawn from $\mathcal{T}_{\text{test}}$. The predictor first produces reward estimates, which are concatenated to the history. The policy then selects actions conditioned on both observed rewards and predictions. The complete implementation procedure is summarized in Algorithm~\ref{algo:ppt}.




\section{Empirical Study}\label{sec:exp}


We compare our proposed PPT algorithm (Algorithm~\ref{algo:ppt}) against DPT in Gaussian multi-armed bandit environments (Definition~\ref{def:gaussbandit}). The goal is to demonstrate that PPT induces more exploratory actions and achieves better online performance, particularly in high-variance environments where effective exploration is crucial for successful learning.

\paragraph{Setup.} 
In all experiments, PPT is tested under varying values of its exploration-weight parameter $\lambda$, which controls the strength of the exploration signal in the training objective. Each setting is denoted as \texttt{PPT\_$\lambda$} in the figures (e.g., \texttt{PPT\_200.0}, \texttt{PPT\_500.0}). To control test environments' difficulty in terms of learning, we vary the variance parameter $\sigma_{\text{test}}^2$, which controls the noise level of reward function. 
We design two types of pretraining datasets to probe complementary aspects of performance. 
\emph{Ideal datasets} are constructed in settings where DPT is known to perform well—using exploratory policies and moderate-length episodes. 
This serves as a sanity check: they ensure that adding an exploration signal in PPT does not sacrifice too much in-distribution performance (as could happen with passive approaches such as injecting constant noise into action probabilities). 
\emph{Tricky datasets}, by contrast, are deliberately long-horizon (posing challenges in data coverage) and expert-biased (encouraging false exploitation). 
These place DPT in unfavorable conditions where offline biases are most likely to induce collapse, allowing us to test whether PPT’s exploration bonus improves robustness in out-of-distribution (OOD) settings. 
The full procedure for generating both datasets is given in the appendix~\ref{appendix:data}. 

For evaluation, we vary $\sigma_{\text{test}}$ and report: 
(i) average suboptimality across the horizon, 
(ii) average cumulative regret across the horizon, 
(iii) the distribution of total regrets across test environments (via smoothed density plots), and 
(iv) the predictor’s online prediction loss (averaged squared $l_2$ distance from true reward vectors).

\begin{figure}[H]
    \centering
    \includegraphics[width=0.48\linewidth]{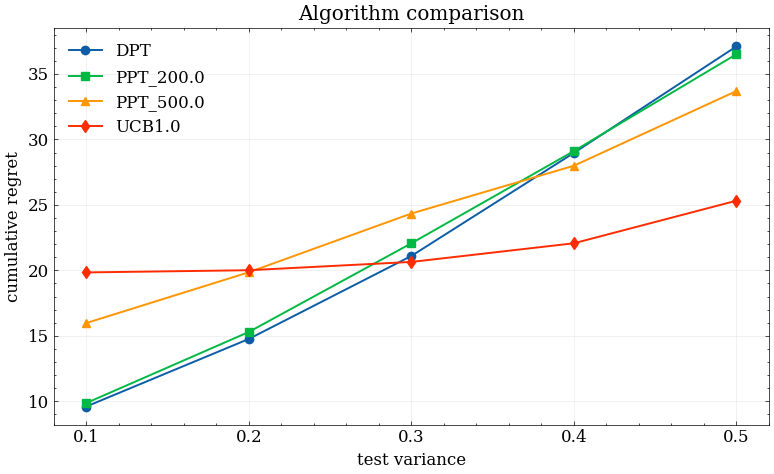}
    \hfill
    \includegraphics[width=0.48\linewidth]{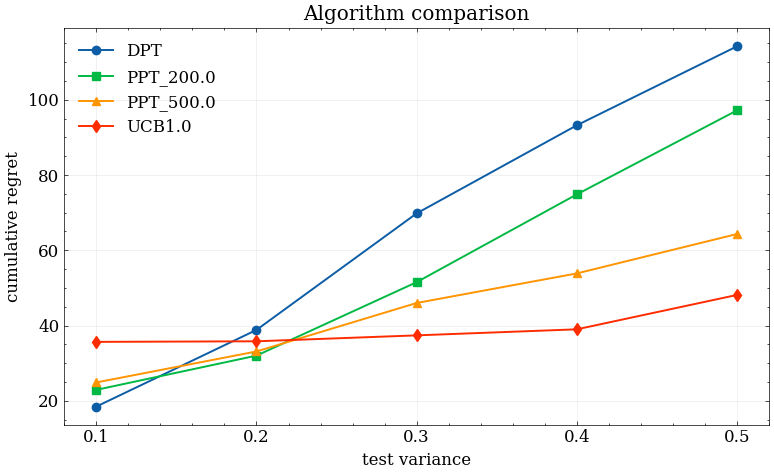}
    \caption{Average regret across increasing $\sigma^2_{\text{test}}$ for (left) ideal and (right) tricky pretraining data. 
    In both settings, PPT shows a lower variance–induced degradation than DPT; the effect is smaller under ideal data and larger under tricky data.}p

    \label{fig:reg-var}
\end{figure}

\paragraph{Performance degradation under increasing variance.}
Fig.~\ref{fig:reg-var} plots the evolution of average regret across the horizon as $\sigma_{\text{test}}$ increases. For both pretraining distributions (ideal and tricky), the DPT curves rise more steeply with variance, while PPT exhibits a slower degradation rate; i.e., the variance-induced gap narrows under PPT in \textit{both} cases. Formally, let

\[
\Delta_{\text{alg}}(\sigma,t) = \text{metric}_{\text{alg}}(\sigma,t) - \text{metric}_{\text{alg}}(\sigma_{0},t), \quad \text{metric} \in \{\text{avg. suboptimality}, \text{avg. regret}\},
\]

where $t$ indexes the horizon and $\sigma_{0}$ is any low-variance baseline (see \ref{appendix:eval} for metric). Empirically, $\Delta_{\text{PPT}}(\sigma,t) < \Delta_{\text{DPT}}(\sigma,t)$ for larger $\sigma$ on both ideal and tricky datasets, with a noticeably smaller effect size under the ideal data (where baseline generalization is already strong) and a pronounced reduction under the tricky data.

\begin{figure}[H]
    \centering
    \includegraphics[width=\linewidth]{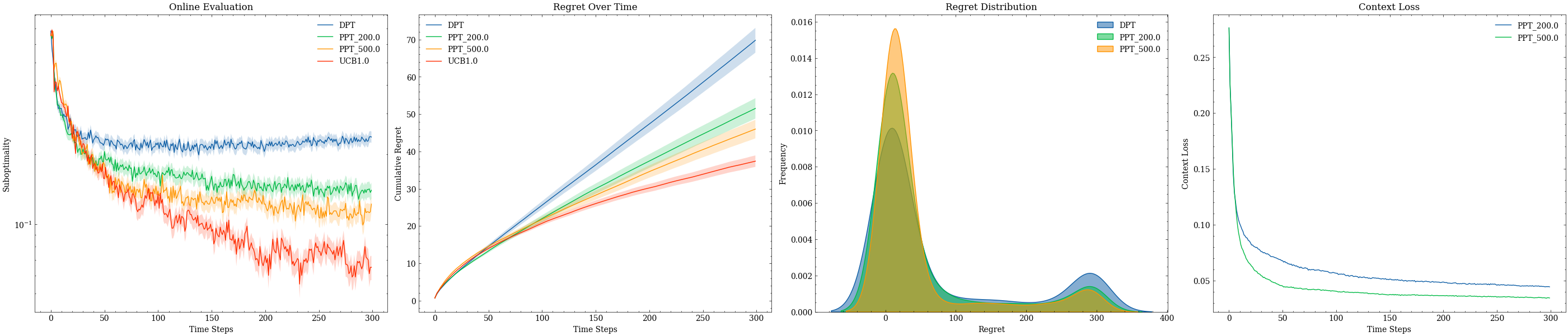}\\
    \includegraphics[width=\linewidth]{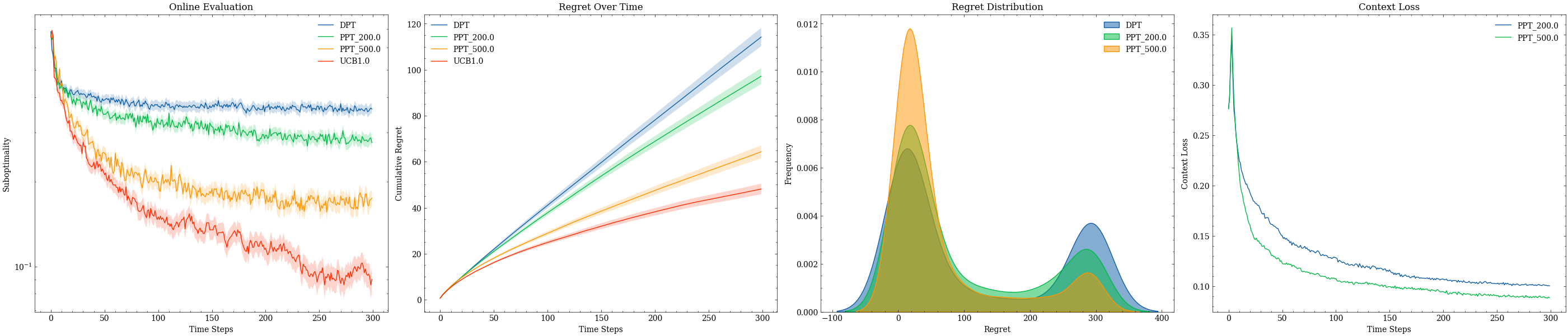}\\

    \includegraphics[width=\linewidth]{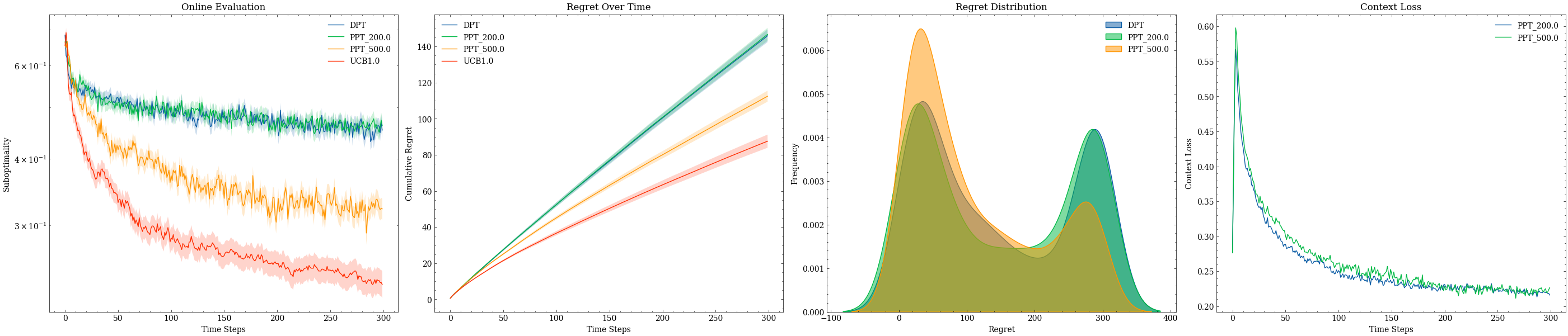}
    \caption{Representative performance of PPT and DPT on test environments with $\sigma^2_{\text{test}}=0.3$ (top), $\sigma^2_{\text{test}}=0.5$ (middle) and  $\sigma^2_{\text{test}}=0.9$ (bottom) using a "tricky"~\ref{def:tky} (more biased towards expert policy, low variance) pretraining data. PPT models exhibit improved generalization relative to DPT. }
    \label{fig:tricky}
\end{figure}

\paragraph{Effect of curiosity coefficient $\lambda$.} 
The choice of $\lambda$ controls the strength of the exploration bias during pretraining. 
Moderate values improve robustness in high-variance test environments while sacrificing little in-distribution performance. 
With $\lambda=0$, PPT reduces to DPT with an extra dimension of input. Empirically, PPT\_0 exhibits very similar behavior to DPT. As we train policy module with increasing $\lambda$, PPT exhibits smaller slope of regret curve in figure~\ref{fig:reg-var}, more stable performance (more concentrated distribution in the third column of figure~\ref{fig:tricky}~\ref{fig:ideal}) and better knowledge about the environments (reflected by lower prediction loss across the horizon in figure~\ref{fig:tricky}~\ref{fig:ideal}).
Excessively large $\lambda$ values may destabilize training, as the policy under-exploits high-reward arms and fails to converge.

\FloatBarrier

\paragraph{Complexity.}
If the predictor and policy models are chosen to have the same architecture and complexity as a baseline DPT policy model, the overall training cost of PPT is approximately $2\times$ that of DPT, since both $\pi_\theta$ and $q_\phi$ are optimized. Note that the training of the predictor module is actually separable from the training of policy model, we can reuse converged predictor models without pretraining for multiple times. If a pretrained predictor is available, the complexity reduces to roughly $1\times$ DPT. At deployment, the predictor contributes an extra forward pass at each step, and inference cost remains comparable to $2\times$ that of DPT.

\section{Discussion}
Our empirical study shows that our proposed PPT algorithm exhibits more exploratory behavior, which consistently reduces performance degradation relative to DPT. 
The effect is robust across both well-exploratory (ideal) and less-exploratory (tricky) pretraining distributions, with stronger improvements when the baseline DPT generalization is weaker. 
Overall, PPT narrows the gap between in-distribution and out-of-distribution environments by incorporating a curiosity-driven regularizer into the pretraining objective. We anticipate that this technique will be effective across a wide range of \textit{state-free} environments with biased or limited pretraining dataset. 

\begin{figure}[htbp]
    \centering
    \includegraphics[width=0.48\linewidth]{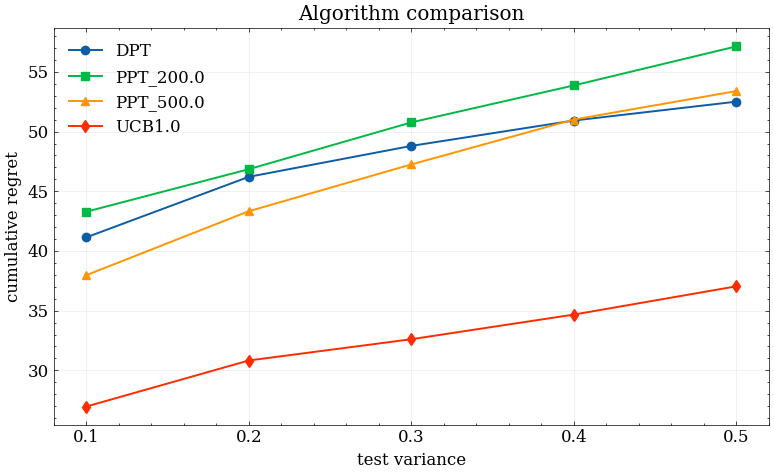}
    \hfill
    \includegraphics[width=0.48\linewidth]{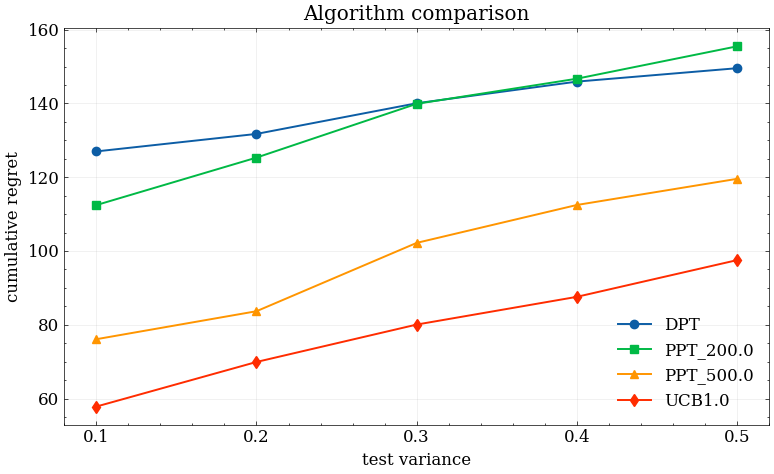}
    \caption{Average regret as test variance $\sigma^2_{\text{test}}$ increases ($\geq0.5$). Left: results with ideal pretraining data. Right: results with tricky pretraining data. As the test environments become more variable, the performance gap diminishes as test variance increases, with PPT converging toward DPT’s regret levels.}

    \label{fig:reg-var_2}
\end{figure}

\paragraph{Limitations.}
The effectiveness of PPT is fundamentally tied to the quality and coverage of the pretraining data. 
When the pretraining distribution lacks sufficient diversity, even curiosity cannot fully close the gap to Bayesian-optimal exploration. 
Thus, PPT should be viewed as an \emph{alternative mechanism} for mitigating dataset bias, not as a complete solution to the exploration problem. As shown in Figure~\ref{fig:reg-var_2}, the advantage of PPT gradually diminishes in highly variable test environments. Moreover, PPT requires access to exact reward information of environments besides offline trajectories, which are more demanding in certain scenarios, though we believe that certain aproximators can be effective replacements.

\paragraph{Future directions.}
Several promising extensions can be envisioned. A natural next step is to move beyond the bandit setting and adapt the framework to in-context reinforcement learning scenarios \textit{with state transitions}. This would require designing curiosity mechanisms that capture uncertainty arising not only from learning the reward function but also from exploring the dynamics of the environment. 

Another direction is to optimize the curiosity algorithm itself. It's possible to relax the reliance on ground-truth arm means by employing approximations such as empirical averages; preliminary evidence suggests that such surrogates may already yield competitive performance (see~\ref{sec:ctx_prox}). Also note that the present training objective is only to maximize the curiosity per arm, it's also meaningful to explore a better functional form for maximizing curiosity.
Furthermore, a particularly important improvement may lie in developing principled strategies for selecting the curiosity coefficient~$\lambda$, or even devising adaptive schemes that adjust it dynamically during training.

\paragraph{}

\bibliographystyle{unsrt}  
\bibliography{references}

\begin{thebibliography}{10}

\bibitem{dt}
Lili Chen, Kevin Lu, Aravind Rajeswaran, Kimin Lee, Aditya Grover, Michael Laskin, Pieter Abbeel, Aravind Srinivas, and Igor Mordatch.
\newblock Decision transformer: Reinforcement learning via sequence modeling, 2021.

\bibitem{laskin2022incontextreinforcementlearningalgorithm}
Michael Laskin, Luyu Wang, Junhyuk Oh, Emilio Parisotto, Stephen Spencer, Richie Steigerwald, DJ~Strouse, Steven Hansen, Angelos Filos, Ethan Brooks, Maxime Gazeau, Himanshu Sahni, Satinder Singh, and Volodymyr Mnih.
\newblock In-context reinforcement learning with algorithm distillation, 2022.

\bibitem{xu2025tackling}
Jiawei Xu, Rui Yang, Shuang Qiu, Feng Luo, Meng Fang, Baoxiang Wang, and Lei Han.
\newblock Tackling data corruption in offline reinforcement learning via sequence modeling.
\newblock In {\em The Thirteenth International Conference on Learning Representations}, 2025.

\bibitem{dpt}
Jonathan~N. Lee, Annie Xie, Aldo Pacchiano, Yash Chandak, Chelsea Finn, Ofir Nachum, and Emma Brunskill.
\newblock Supervised pretraining can learn in-context reinforcement learning, 2023.

\bibitem{mukherjee2025pretrainingdecisiontransformersreward}
Subhojyoti Mukherjee, Josiah~P. Hanna, Qiaomin Xie, and Robert Nowak.
\newblock Pretraining decision transformers with reward prediction for in-context multi-task structured bandit learning, 2025.

\bibitem{rietz2025enhancingpretraineddecisiontransformers}
Finn Rietz, Oleg Smirnov, Sara Karimi, and Lele Cao.
\newblock Enhancing pre-trained decision transformers with prompt-tuning bandits, 2025.

\bibitem{hu2024qvalueregularizedtransformeroffline}
Shengchao Hu, Ziqing Fan, Chaoqin Huang, Li~Shen, Ya~Zhang, Yanfeng Wang, and Dacheng Tao.
\newblock Q-value regularized transformer for offline reinforcement learning, 2024.

\bibitem{lin2024transformers}
Licong Lin, Yu~Bai, and Song Mei.
\newblock Transformers as decision makers: Provable in-context reinforcement learning via supervised pretraining.
\newblock In {\em The Twelfth International Conference on Learning Representations}, 2024.

\bibitem{wang2025understanding}
Hanzhao Wang, Yu~Pan, Fupeng Sun, Shang Liu, KALYAN~TEJA TALLURI, Guanting Chen, and Xiaocheng Li.
\newblock Understanding the training and generalization of pretrained transformer for sequential decision making, 2025.

\bibitem{goddard2025incontextlearninggeneralizetask}
Chase Goddard, Lindsay~M. Smith, Vudtiwat Ngampruetikorn, and David~J. Schwab.
\newblock When can in-context learning generalize out of task distribution?, 2025.

\bibitem{pathak2017curiositydrivenexplorationselfsupervisedprediction}
Deepak Pathak, Pulkit Agrawal, Alexei~A. Efros, and Trevor Darrell.
\newblock Curiosity-driven exploration by self-supervised prediction, 2017.

\bibitem{burda2018explorationrandomnetworkdistillation}
Yuri Burda, Harrison Edwards, Amos Storkey, and Oleg Klimov.
\newblock Exploration by random network distillation, 2018.

\end{thebibliography}

\appendix
\section{Experimental Details}
\subsection{Environments and Datasets}\label{appendix:data}

Both pretraining and test datasets are generated by first defining a distribution over bandit environments and then sampling independently.  

\begin{definition}[Gaussian Bandit]\label{def:gaussbandit}
A Gaussian bandit environment is represented as the tuple $(\mathcal{A}, \mu, \sigma, n)$, where the mean reward vector $\mu \in \mathbb{R}^{|\mathcal{A}|}$ and standard deviation $\sigma$ specify the reward distribution as $R(a_j) = \mathcal{N}(\mu_j, \sigma^2)$ for each arm $a_j \in \mathcal{A}$.
\end{definition}

\subsubsection{Training}
We pretrain on Gaussian bandits sampled from  
\begin{equation}\label{eqn: t_pre}
    \mathcal{T}_{\text{pre}} = (\mathcal{A} = [3], \sigma^2 \sim \text{Unif}(I), \mu \sim \text{Unif}[0,1]^3, n = n_{\text{train}})
\end{equation}
where $I = [0.1, 1.0]$ by default. Each episode is collected using the data-collection policy  
\begin{equation}\label{eqn: d_pre}
    \mathcal{D}_{\text{pre}}(a_{j+1}|D_j, \tau) = w \cdot e_{i(\tau)} + (1 - w) \cdot p_j, \quad p_j \sim \text{Dir}(1^{|\mathcal{A}|})
\end{equation}
where $w$ controls the bias toward expert actions, and $i(\tau)$ is the expert action index.

We consider two types of pretraining datasets:

\begin{definition}[Ideal dataset]\label{def:idl}
$w = 0.2$, moderate bias toward expert trajectories, episode variance sampled from $\sigma^2 \sim \text{Unif}[0.1, 1.0]$.
\end{definition}

\begin{definition}[Tricky dataset]\label{def:tky}
$w = 0.8$, high bias toward expert trajectories, low variance $\sigma^2 = 0.1$ to trick into early exploitation.
\end{definition}

\subsubsection{Evaluation}
For evaluation, we define a test distribution $\mathcal{T}_{\text{test}}$ and sample 1000 independent environments from it:
\begin{equation}
    \mathcal{T}_{\text{test}} = (\mathcal{A} = [3], \sigma = \sigma_{\text{test}}, \mu \sim \text{Unif}[0,1]^3, n = n_{\text{train}})
\end{equation}
This ensures robust results that are largely immune to environment stochasticity, while capturing algorithmic differences.

\subsection{Model Hyper-parameters and Evaluation}\label{appendix:eval}
We train PPT and DPT on the same datasets to ensure fair comparison. All experiments were run on a single NVIDIA RTX 4090 GPU. Training times were approximately 1.5 hours for PPT and less than 1 hour for DPT. Both PPT and DPT models use HuggingFace's Transformers library, with training implemented in PyTorch. We use the AdamW optimizer with weight decay 1e-4, learning rate 1e-3, and batch-size 256. We use an embedding size of 32 and 4 layers for all models, constrained by computational resources, though performance of both PPT and DPT models are scalable. 

For evaluation, each group of 1000 test environments $\{\tau^{(i)}\}_{i \in [1000]} \sim \mathcal{T}_{\text{test}}$ is used to compare the following algorithms:

\paragraph{PPT-$\lambda$.} Prediction Pretrained Transformer with weight $\lambda$. In our case, A default tuning range of $\lambda$ for stable performance and training is $[100, 500]$.
\paragraph{DPT.}
Original Decision-Pretrained Transformer from~\cite{dpt}.  
\paragraph{Upper Confidence Bound (UCB).}
Upper Confidence Bound policy chooses action by:
\begin{equation}
    a_t = \arg\max_a \left[\hat{\mu}_a + \beta \sqrt{\frac{1}{n_a}} \right]
\end{equation}
where \( n_a \) denotes the number of times arm \( a \) has been selected at step $t$ and $\hat{\mu}_a$ denotes the empirical mean so far for arm $a$. We found that $\beta = 1.0$ performs well in the wide range of test environments we consider and thus take it as the default constant. We consider this as the baseline for state-of-art performance for model-free algorithms on our test environments.

The following metrics correspond to those used in the figures of the main text (e.g., Fig.~\ref{fig:tricky}):

\begin{align}
\texttt{avg.suboptimality}(t) &= \frac{1}{1000}\sum_{i = 1}^{1000} [\mu^{*(i)} - \langle a^{(i)}_t, \mu^{(i)} \rangle] 
\end{align}

\begin{align}
\texttt{avg.regret}(t) &= \frac{1}{1000}\sum_{i = 1}^{1000} \sum_{j = 1}^{t} [\mu^{*(i)} - \langle a^{(j)}_t, \mu^{(i)} \rangle] 
\end{align}

where $\mu^{(i)} \in \mathbb{R}^{|\mathcal{A}|}$ is the mean reward vector of environment $\tau^{(i)}$, $\mu^{*(i)} \in \mathbb{R}$ is the mean reward of the best arm, and $a_j^{(i)} \in \Delta(\mathcal{A})$ is the arm probability of the algorithm at time step $j$. For deterministic policies, $a_j^{(i)}$ is a one-hot vector.

\subsection{Extra Experimental Results}

\subsubsection{Training Dynamics}

\begin{figure}[t]
    \centering
    \vspace{0em} 
    \includegraphics[width=0.45\linewidth]{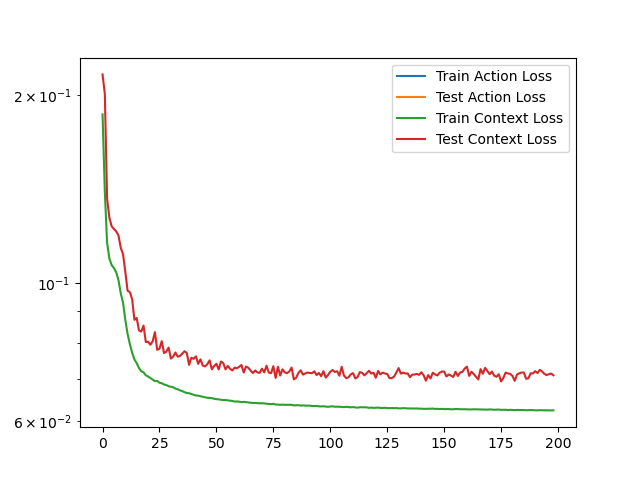}
    \hfill \includegraphics[width=0.45\linewidth]{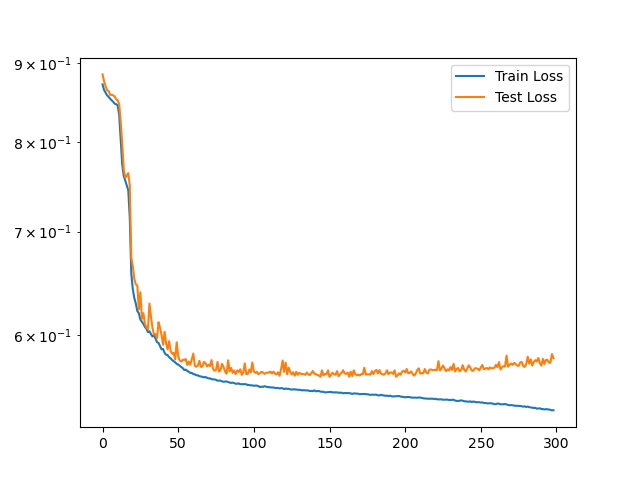}
    \caption{Policy loss shares similar dynamics with predictor loss during training.}
    \label{fig:loss_32}
    \vspace{0em}
\end{figure}

We report that training dynamics of both policy and predictor model are governed primarily by dataset characteristics and model complexity.

During the optimization of the predictor module and 
The predictor model in PPT exhibits identical training dynamics to DPT under matched latent dimensions. 

Following stabilization of the predictor module, PPT’s action-loss trajectory converges quantitatively with DPT’s behavioral profile.

The observed parity suggests that PPT’s objective of learning latent environment structures and DPT’s goal of optimizing per-round actions represent closely related problem spaces in terms of their optimization landscapes.

\begin{figure}[t]
  \centering
  \begin{subfigure}{0.329\textwidth}
    \centering
    \includegraphics[width=\textwidth]{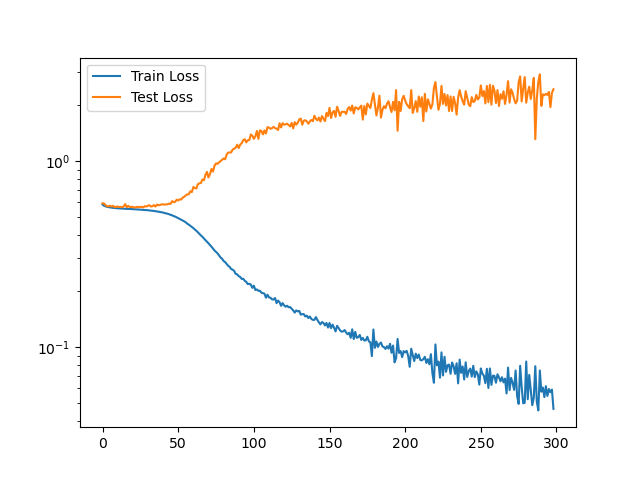}
    \caption{DPT}
  \end{subfigure} 
  \hfill
  \begin{subfigure}{0.329\textwidth}
    \centering
    \includegraphics[width=\textwidth]{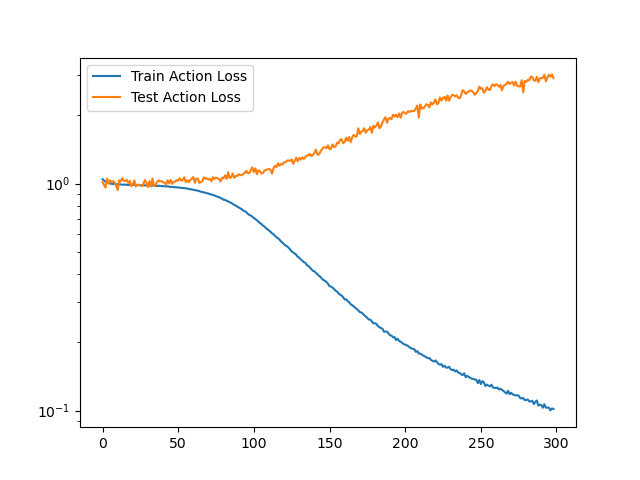}
    \caption{PPT with $\lambda = 100$}
  \end{subfigure}
  \hfill
  \begin{subfigure}{0.329\textwidth}
    \centering
    \includegraphics[width=\textwidth]{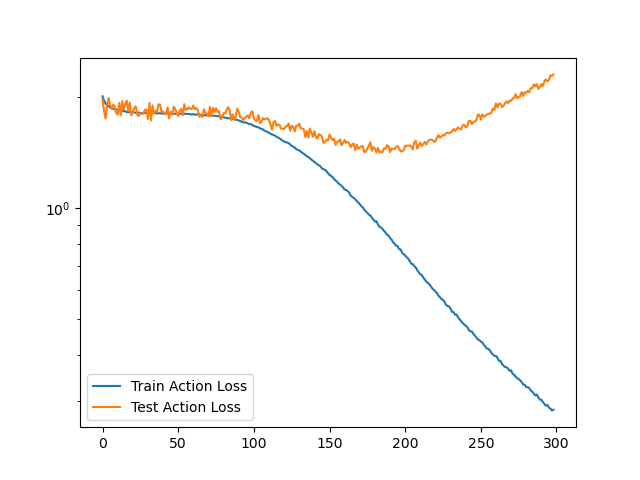}
    \caption{PPT with $\lambda = 500$}
  \end{subfigure}
  \caption{Training dynamics of DPT and PPT (embedding size 256, 6 layers) on the ideal dataset.}
  \label{fig:loss_256}
\end{figure}
\FloatBarrier

\subsubsection{Training with proxy context instead of ground truth}\label{sec:ctx_prox}

We also tested PPT under a modified setup where the context signal is not the ground-truth mean reward vector but a \emph{per-trajectory mean estimator}~\ref{eqn:ctx_prox}. 
The approximation of the true mean reward vector is given by:
\begin{align}\label{eqn:ctx_prox}
    \hat{c}^* \in \mathbb{R}^{|\mathcal{A}|}, \quad \hat{c}^*[i] = \frac{1}{n} \sum_{j = 1}^n r_j \cdot 1_{\{a_j = i\}}
\end{align}

which is computed for each full episode.

\begin{figure}[H]
    \begin{subfigure}{0.45\textwidth}
    \centering
    \includegraphics[width=\linewidth]{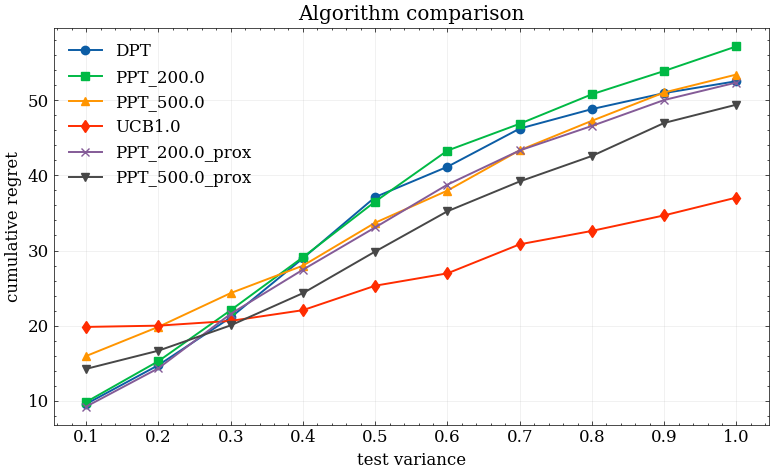}
    \caption{Trained on ideal dataset}
    \end{subfigure}
    \hfill
    \begin{subfigure}{0.45\textwidth}
    \centering
    \includegraphics[width=\linewidth]{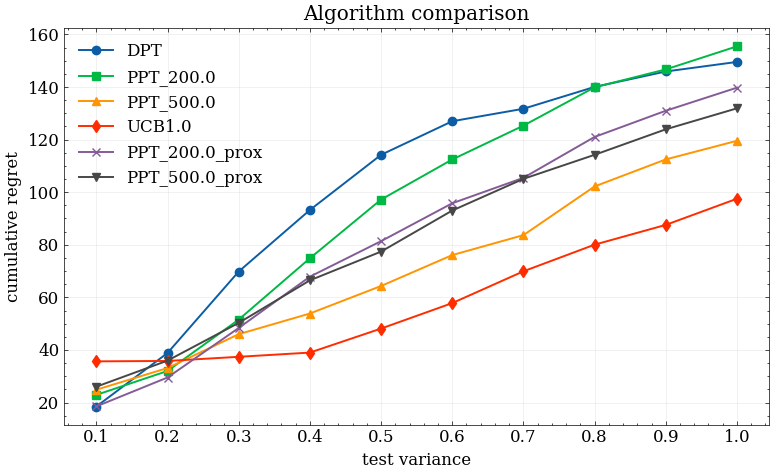}
    \caption{Trained on tricky dataset}
    \end{subfigure}
    \caption{PPT trained on proxy context achieve comparable regret performance on wide range of test variances.}
    \label{fig:prox}
\end{figure}

Interestingly, models pretrained with such proxy contexts achieved performance comparable to, and in some cases slightly better than, those trained with ground-truth rewards (see Fig.~\ref{fig:prox}). 
The results suggest that proxy contexts can still provide a useful exploratory bias for the policy model, even without direct access to the true mean rewards. 

We emphasize that this is only a preliminary analysis: the experiments were restricted to the \emph{ideal} and \emph{tricky} datasets introduced earlier, without exploring broader varieties of pretraining distributions. More systematic investigation of proxy contexts is left as future work.

\subsubsection{Extra Plots}

We presented plots that were not used in previous sections here.

\begin{figure}[H]
    \centering
    \includegraphics[width=\linewidth]{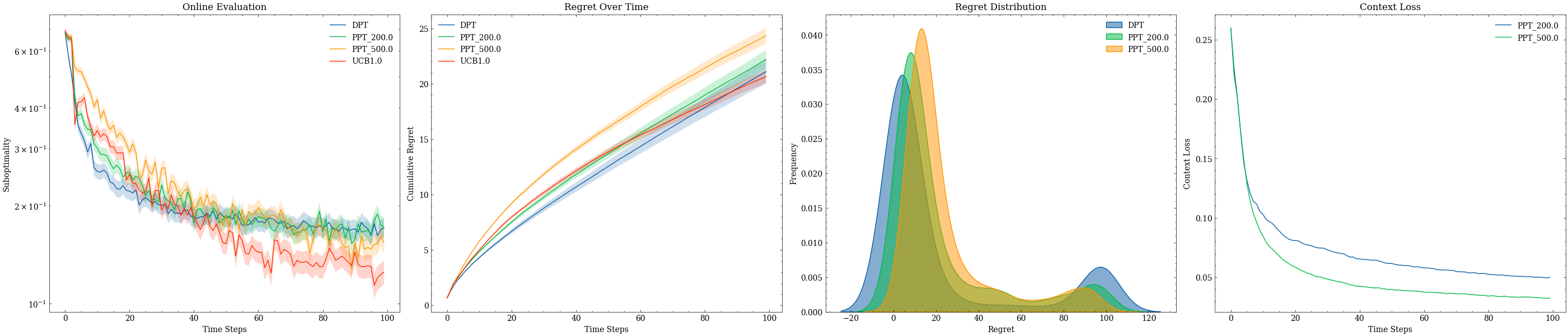}\\
    \includegraphics[width=\linewidth]{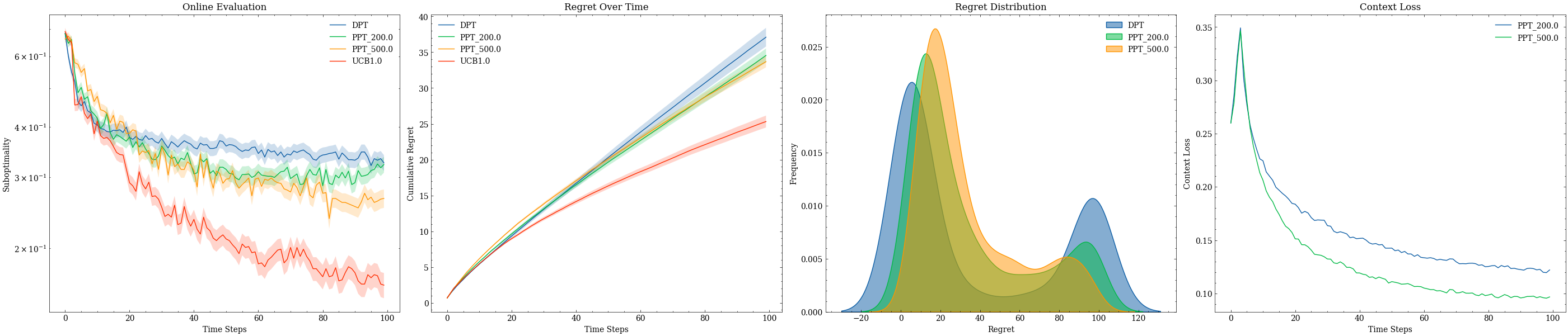}\\
    \includegraphics[width=\linewidth]{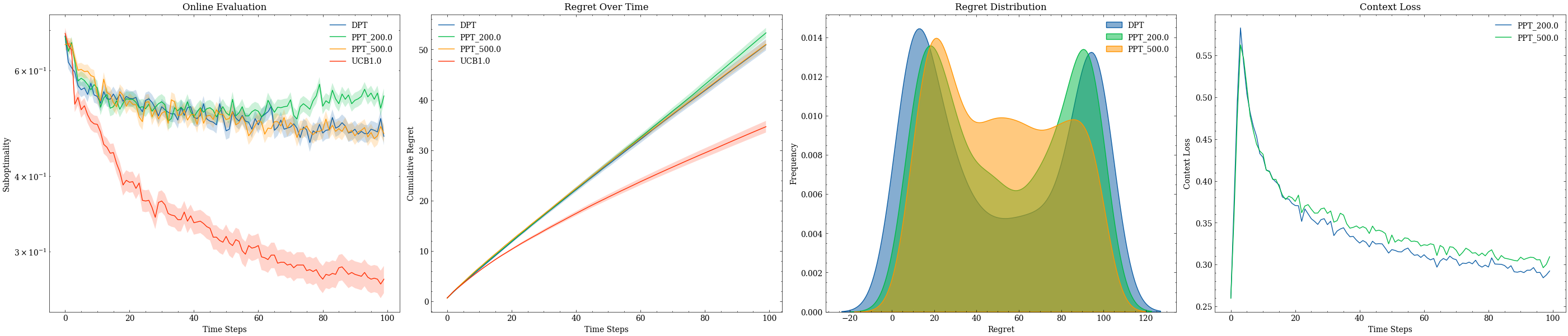}\\
    
    \caption{Performance of PPT and DPT on test environments with $\sigma^2_{\text{test}}=0.3$ (top), $\sigma^2_{\text{test}}=0.5$ (middle) and  $\sigma^2_{\text{test}}=0.9$ (bottom) using ideal~\ref{def:idl} pretraining data.}
    \label{fig:ideal}
\end{figure}

\begin{figure}[H]
    \centering
    \includegraphics[width=\linewidth]{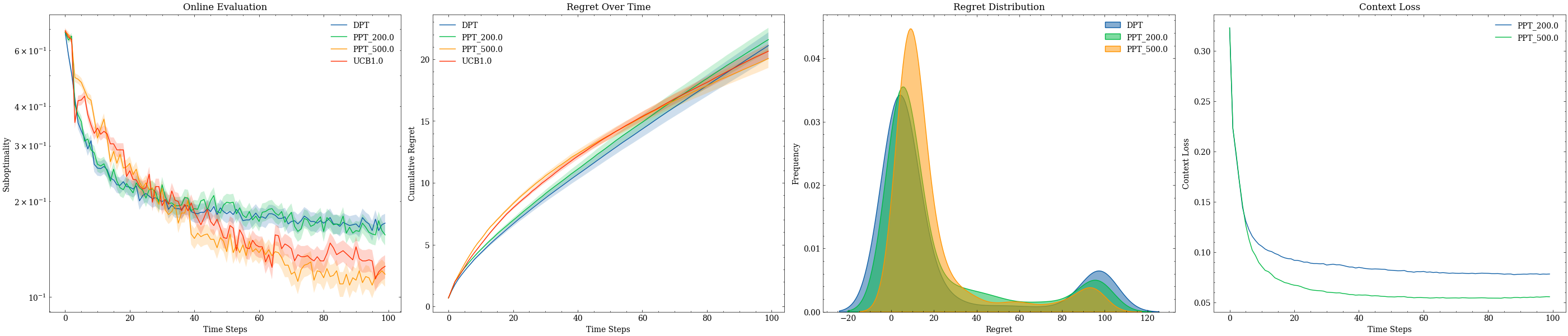}
    \includegraphics[width=\linewidth]{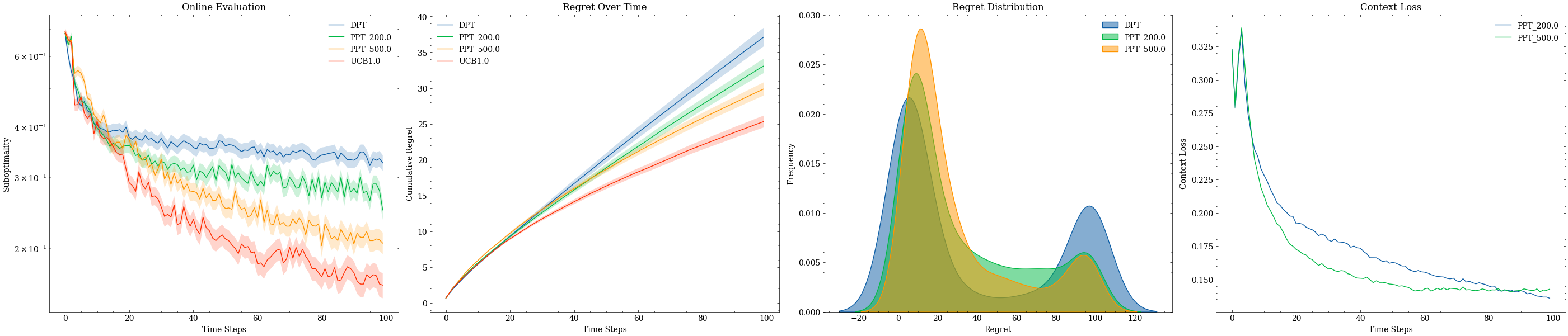}
    \includegraphics[width=\linewidth]{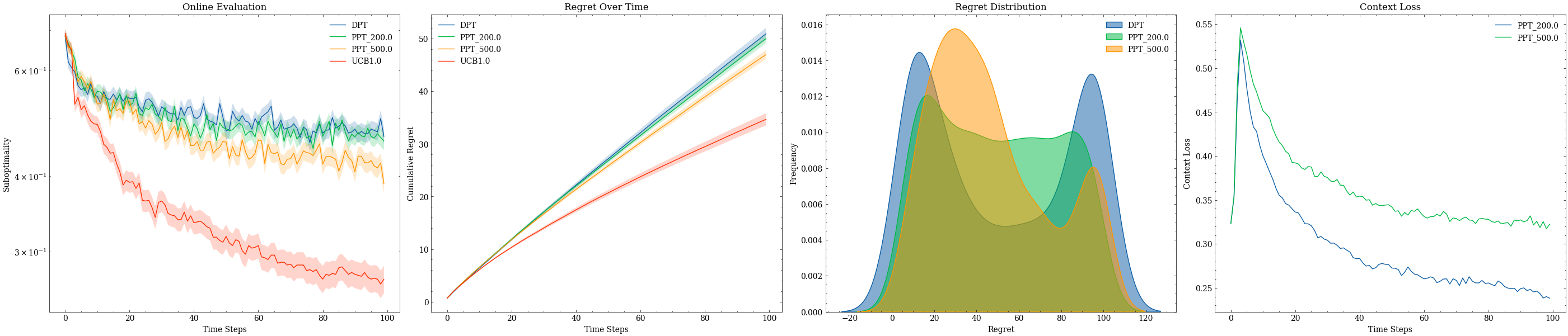}
    \caption{Rollouts for PPT trained with proxy context and ideal~\ref{def:idl} dataset on test environments with $\sigma^2_{\text{test}}=0.3$ (top), $\sigma^2_{\text{test}}=0.5$ (middle) and  $\sigma^2_{\text{test}}=0.9$ (bottom).}
    \label{fig:ideal_prox}
\end{figure}

\begin{figure}[H]
    \centering
    \includegraphics[width=\linewidth]{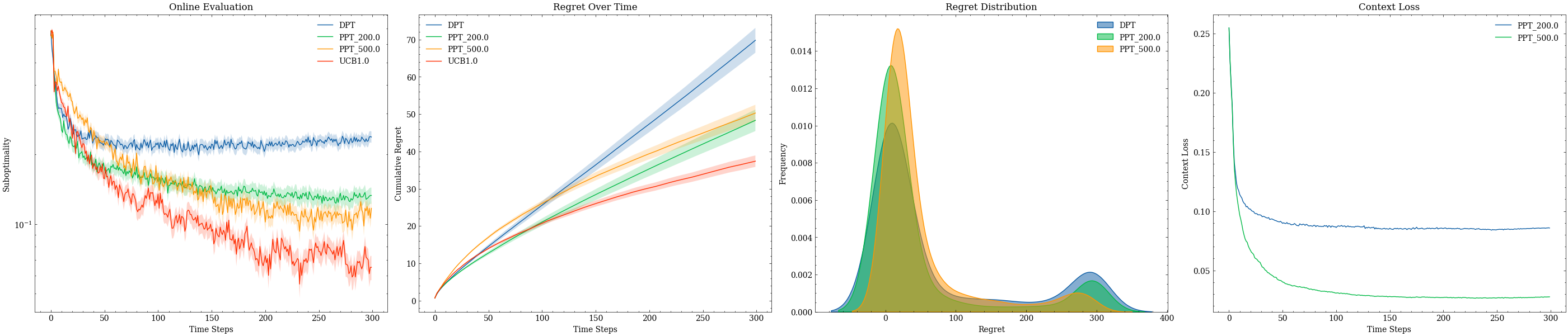}
    \includegraphics[width=\linewidth]{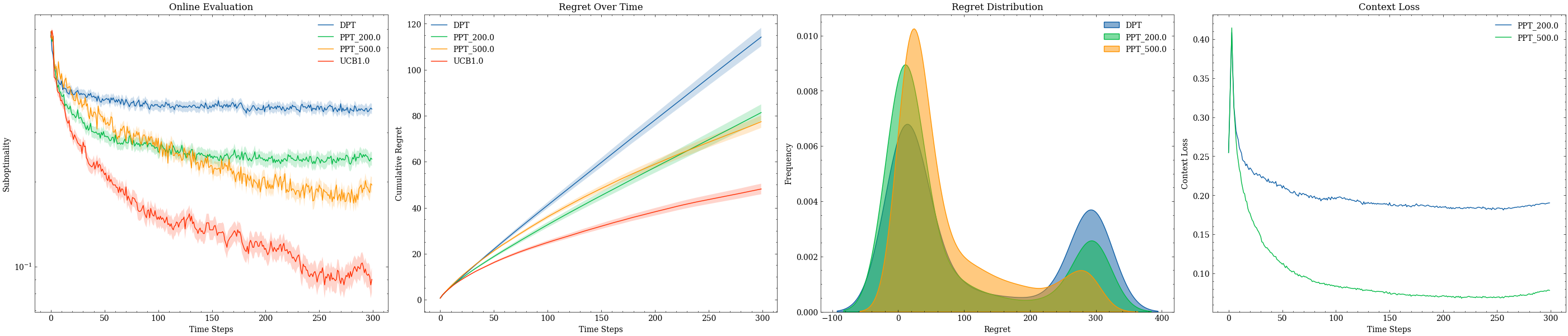}
    \includegraphics[width=\linewidth]{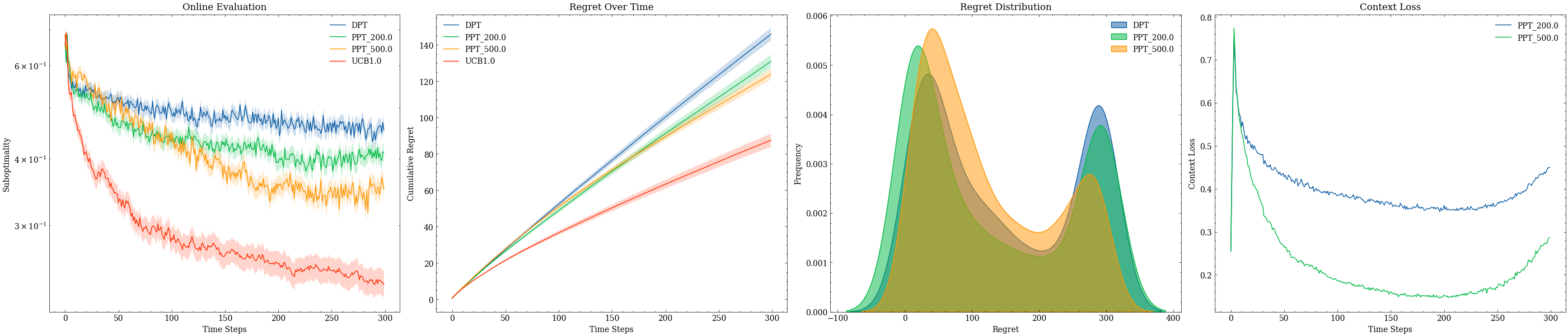}
    \caption{Rollouts for PPT trained with proxy context and tricky~\ref{def:tky} dataset on test environments with $\sigma^2_{\text{test}}=0.3$ (top), $\sigma^2_{\text{test}}=0.5$ (middle) and  $\sigma^2_{\text{test}}=0.9$ (bottom).}
    \label{fig:tricky_prox}
\end{figure}

\end{document}